# Multidimensional Scaling for Gene Sequence Data with Autoencoders

Pulasthi Wickramasinghe
SICE
Indiana University
Bloomington, IN USA
pswickra@iu.edu

Geoffrey Fox
SICE
Indiana University
Bloomington, IN USA
gcf@indiana.edu

*Abstract*— Multidimensional scaling of gene sequence data has long played a vital role in analysing gene sequence data to identify clusters and patterns. However the computation complexities and memory requirements of state-of-the-art dimensional scaling algorithms make it infeasible to scale to large datasets. In this paper we present an autoencoder-based dimensional reduction model which can easily scale to datasets containing millions of gene sequences, while attaining results comparable to state-of-the-art MDS algorithms with minimal resource requirements. The model also supports out-of-sample data points with a 99.5%+ accuracy based on our experiments. The proposed model is evaluated against DAMDS with a real world fungi gene sequence dataset. The presented results showcase the effectiveness of the autoencoder-based dimension reduction model and its advantages.

## I. Introduction

Deep learning has been emerging as a solution to many machine learning applications and is shown to have strong performance and versatility in many areas, which has led to its adaptation in both industry and academia for machine learning applications. With this expansive application in many fields, more and more deep learning models have been introduced to tackle specific sub-problem areas. The increased availability of data has also led to the use of deep learning models for various applications. This is because with more data readily available, larger and more complex networks can be created and trained to produce more accurate results. Thanks to the exponential growth of data in the modern world, this trend will only continue to accelerate. One of the most important aspects of deep learning that makes it such a versatile solution is that networks are able to learn and encode patterns and structures present within the input data very efficiently. Thus this learned information can be used to categorize or characterize new input data.

Autoencoders are a widely used unsupervised neural network model which learn how to represent a dataset using a reduced representation and then reconstruct that data back from the reduced version to the original representation, essentially multidimensional scaling. This paper presents a model which utilizes the natural dimension reduction capability in autoencoders as a solution to dimension reduction in gene sequence data.

Multidimensional scaling (MDS) refers to a set of algorithms and methods that are used to convert high dimensional data into lower dimensions so they can be visualized and analysed. MDS is a well-researched area, especially in machine learning, and there are various linear and non-linear solutions available in research literature [1], [2], [3]. This is important because humans are not able to clearly think about data in higher dimensions, so scaling them down to 2 or 3 dimensions allows data to be visualized to find patterns and structures in the data. One of the most important properties of multidimensional scaling is that it should keep as much information from the data points as possible while projecting them down to lower dimensions. There are various MDS algorithms used in modern applications. In [2], the authors discuss such methods. Additionally, new MDS methods such as T-SN [4] and UMAP [5] have gained more popularity over recent years. DAMDS [6] is another popular scalable MDS algorithm used for large datasets. As with many other machine learning domains, deep learning can be applied to MDS. While the literature on using deep learning for dimension reduction is small at the moment, research is being done to achieve better results using the latest such technologies.

Scaling data using MDS techniques to 2 or 3 dimensions plays an important role in analysing biological data such as gene sequences or biological images. This allows researchers to identify clusters and patterns within large datasets which would otherwise be incomprehensible due to the high dimensionality of the data. With advances in bio-sequencing techniques, more and more raw gene sequence data is readily available to be analysed. While we limit the scope of this paper to biological gene sequence data, approaches along the same lines as proposed herein might be adaptable for various other forms of biological data, such as biological image datasets.

A major obstacle in using MDS algorithms for large datasets that contain hundreds of thousands, if not millions, of data-points is the computation complexity and memory requirements of these algorithms, which are typically $O(N^2)$ where $N$ is the number of data points. This makes it difficult for even parallelized implementations of MDS algorithms to scale for large datasets. In addition, MDS algorithms do not have straightforward extensions or methods to calculate the embeddings of out-of-sample data points without performing the complete calculation from scratch.

In this paper we propose an autoencoder-based solution for multidimensional scaling which would be able to address both the scalability and out-of-sample data point issues with regard to gene sequence data. While the scope of this paper is limited to gene sequence data, many of the concepts should be applicable to other areas as well. To this end, we present experiments and results showing the effectiveness and performance of the proposed autoencoder-based models using a large real gene sequence dataset.

The outline of the paper is as follows. Section II provides a brief introduction into MDS and touches upon the current state-of-the-art options in MDS. In Section III we provide an overview of autoencoders and why they fit the problem of MDS so well. Section IV explains how autoencoders can be used for dimension reduction and the main challenges that need to be addressed to adapt for gene sequence data. In

Section V we present two solutions on how gene sequence data can be adapted to work with autoencoders and discuss the merits of each approach. Finally, in Section VI we present experimental results to show how effective and accurate the proposed approaches are for dimension reduction of gene sequence data.

## II. MULTIDIMENSIONAL SCALING

Multidimensional scaling [7], [8] is a term used to broadly classify techniques and algorithms that can represent higher dimensional data in lower dimensional space such as 2 or 3 dimensions. The goal of MDS algorithms is to keep information loss to a minimum when projecting data into lower dimensions. This is done based on pairwise distance information of the data. In essence, MDS is a non-linear optimization problem which tries to optimize the mapping in the target dimension based on the original pairwise distance information. The use of pairwise distance information allows MDS to be used with data such as biological gene sequence data, which typically cannot be represented using a set of feature vectors but does have pairwise distances. This allows MDS to apply to a broader set of datasets, unlike other dimension reduction methods such as PCA, SOM, etc. MDS algorithms have a high computation complexity and memory requirement, which is normally $O(N^2)$ where $N$ is the number of data points. The memory requirement is evident by the fact that they take in an *NxN* input distance matrix. Efficient parallel implementations [9] and approximation-based implementations [10], [11] of MDS have provided the ability for MDS algorithms to be applied to larger datasets, yet they are still limited by memory availability of the cluster. For the purpose of this paper, we employ MPI-based DAMDS [12], which is a high performance implementation of the WDA-SMACOF [6] algorithm.

### A. Input Data Preparation

Gene sequence data in its raw form cannot be used as inputs to MDS. MDS algorithms take in a pairwise distance matrix as input, therefore the raw gene sequence data needs to be processed to generate the corresponding distance matrix. To this end, we use Smith-Waterman (SW) [13] algorithm to calculate pairwise distance data. It is important to note that the distance matrix generation step itself is quite compute intensive since $N^2$ distances need to be calculated. The Smith-Waterman algorithm itself also has a complexity of $O(ML)$ where $M$ and $L$ are the lengths of the gene sequences being aligned. An important point regarding the use of pairwise distance calculated through a sequence alignment algorithm such as SW is that it embeds some form of biological similarity information directly in the distance values.

### B. Out-of-sample data

Out-of-sample data refers to data points that were not contained in the original dataset but need to be projected into the same lower dimensional space. For example, these might be a new batch of gene sequences that were not initially added in the dataset but still need to be made part of the analysis. In such instances it is important to be able to embed the out-of-sample data points without having to completely recalculate the embedding for the whole dataset. Even though there is typically no direct extension for MDS algorithms to support out-of-sample data, work has been done to provide such extensions in [14], [15]. However we believe the autoencoders-based model introduced in this paper provides

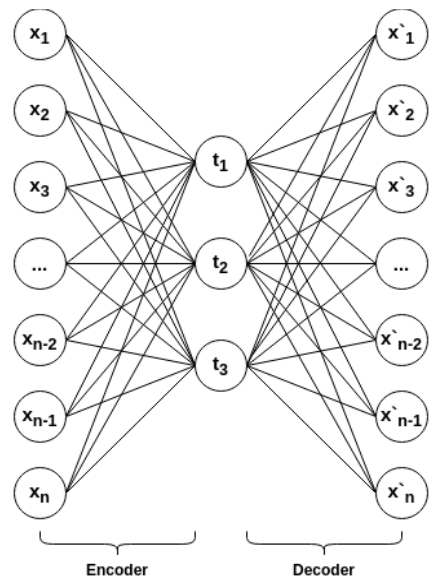

Fig. 1. The structure of an autoencoder

a much simpler and more straightforward solution for out-of-sample data points.

## III. AUTOENCODER

Based on how the various layers of the network are designed, neural networks can be classified into various categories such as convolutional neural network, recurrent neural network, etc. Autoencoders are one such neural network model. An autoencoder neural network is an unsupervised learning model since it does not require labeled data to train the network. The layers in the autoencoder are designed in such a way that it contains a bottleneck layer, as shown in Fig. 1. The input layer and the final output layer of an autoencoder have an equal number of units. Typically a network of the autoencoder is broken down into two major segments: the section from the input layer up to the bottleneck layer is known as the Encoder, and the section from the bottleneck layer to the output layer is known as the Decoder. The aim of the network is to take in the input, propagate it through the network, and try to reconstruct the input at the output layer as accurately as possible. The loss function of an autoencoder is typically some measurement of the difference between input and output vectors. Since the network has a bottleneck in the middle, in order to minimize the loss, the network needs to learn how to embed the input data using a lower number of units. This means that autoencoders essentially perform dimensional reduction to learn the most efficient representation in a lower dimension so it can reconstruct data from the reduced representation. Fig. 1 shows a network with a single hidden layer, but autoencoders can be designed with several hidden layers as needed.

## IV. MDS WITH AUTOENCODER

Once the basic mechanics of an autoencoder are realized, the way in which they are used as a mechanism to perform dimension reduction can be understood clearly. Looking at the function performed by the autoencoder in Fig. 1 from a multidimensional reduction point of view, the network takes in an input with "$n$" dimensions and reduces it at the hidden layer to a 3-dimensional data point. Once the network is trained, the decoder section of the network is discarded and the trained weights can be fixed. During the evaluation phase, input given into the network will output a data point with "*d*" number of dimensions, where "*d*" is the number of units in the bottleneck hidden layer (which is the output layer since the decoder was discarded). In [16] the authors evaluate how autoencoder-based dimension reduction compares to other forms of dimension reduction such as PCA, Isomap, etc. They find that the autoencoder-based approach does produce comparable results for both synthetic and real world data, however the experiments were done on rather small datasets.

The main challenge with adapting autoencoders to perform dimension reduction on gene sequence data is determining how the input data, which are gene sequences, can be modeled as inputs to the network. Autoencoders expect an input as a fixed length vector, however gene sequences in a dataset are not typically of equal length and consist of characters. Autoencoders expect inputs to be of numerical value. In Section V we present a few solutions to address this issue, and in Section VI we evaluate the effectiveness of the proposed approaches.

## V. ADAPTING GENE SEQUENCES FOR AUTOENCODERS

In order to use gene sequence data as input data for the autoencoder, the raw sequences need to be transformed into a format that can be consumed by the autoencoder. To this end, the gene sequence, which consists of a sequence of characters, must be converted into a representative numerical vector.

### A. Encoding gene sequences

Choosing the correct encoding to convert the gene sequence into a fixed-size vector is important since some encoding may provide false signals to the network, resulting in poor performing networks. The two most common methods used to encode sequence data into numeric vectors are "one hot encoding" and "ordinal encoding."

#### 1) One Hot Encoding

In One Hot Encoding (OHE), each character in the sequence is replaced by a vector. The length of the vector is equal to the number of unique characters in the sequence, which is 4 in RNA sequences, and the values in the vector other than the location that corresponds to the current character would be zero. All these vectors are then concatenated to create a single vector that represents the whole sequence. For example, an RNA sequence *ATGC* would have the encoding shown below. Since gene sequences in the same dataset may have different lengths, to keep the size of the input vectors equal, vectors need to be padded with "0" when needed.
*Ex: ATGC - [0,0,0,1], [0,0,1,0], [0,1,0,0], [1,0,0,0]*

#### 2) Ordinal Encoding

In ordinal encoding, instead of a vector, each character is given a numeric value. The sequence encoding replaces each character with the corresponding numeric value. As with OHE, to make vector lengths equal, "0" must be added as padding values when needed. Taking RNA sequences as an example, characters "*A,T,G,C*" can be assigned values "0.25, 0.5, 0.75, 1.0" respectively. This would mean the encoded result of the sequence "GGTAC" would be as follows.
*Ex: GGTAC - [0.75, 0.75, 0.5, 0.25, 1.0]*

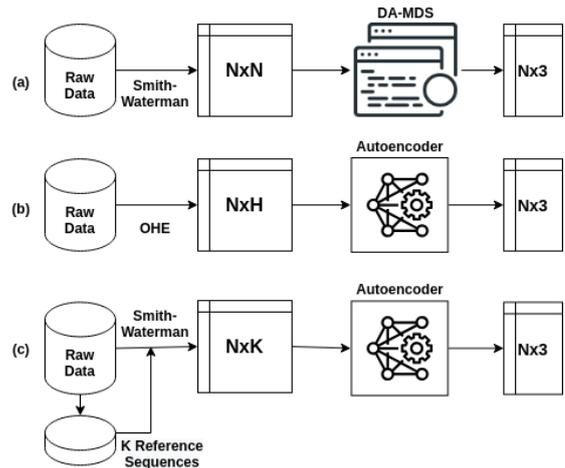

Fig. 2. Evaluated dimension reduction approaches. (a) DAMDS. (b) Autoencoder with OHE, "H" is the length of the vector once it is encoded with OHE, (c) Autoencoder with reference sequences, "K" is the number of reference sequences

One hot encoding has a drawback in that for sequence data with a large alphabet, it would create very large encoding since the length of the vector is the size of the alphabet for each character in the sequence. While ordinal encoding does not have this problem, it has a much more acute issue when applied for neural networks. Since different numeric values are given to different characters of the alphabet, the network tends to give more weight to higher numeric values, which would result in the trained network favouring certain characters of the alphabet rather than others. In the context of gene sequences, OHE seems to be the most suitable and popular encoding method. Since the alphabet of the gene sequence is limited (4 letters in RNA and 25 if converted to amino acids), OHE does not have a large impact when encoding gene sequence data. The argument to use OHE to convert gene sequences to vectors is further validated by its use in the research literature, as shown in [17], [18]. Fig. 2 (b) illustrates how OHE is used to generate input vectors for one of the three dimension reduction approaches evaluated in Section VI.

### B. Pairwise Distance

As mentioned in Section II, pairwise distances calculated through gene sequence alignment algorithms have the added advantage of incorporating biological similarity information in the distance calculation. The main drawback of using OHE to encode input data is that it does not embed any biological distance data in the generated embedding vector, which is used as the input to the neural network. However, directly using the distance matrix as with MDS is not practical since the distance matrix generated is an *NxN* matrix, and thus the input vector for each input data will be of length *N* where *N* is the number of data points in the dataset. This would mean that the input layer of the autoencoder would consist of *N* units and would have to change with the size of the dataset.

As a solution to this, we propose a sampling-based solution which allows biological distance information to be embedded in the input while keeping the length of the input length fixed. For a given dataset with *N* sequences, *K* random sequences are selected as reference sequences. Next the pairwise distances between each sequence and the *K* reference sequences are calculated, which produces a vector to length *K* to be used as the input for the autoencoder. Choosing a representative *K* depends on the dataset, but it needs to be sufficiently large to make sure the loss of information is minimal. For example, if the goal is to identify clusters within the dataset, a logical estimate of *K* would have to be "*10xC*" or "*20xC*," where "*C*" is a rough estimate of the number of clusters in the dataset. Keeping *K* as large as possible based on memory and computation capacity restriction would make sure the information loss is minimal, but it would also increase the size of the autoencoder network. Results discussed in Section VI show that even small *K* produces good results. Section VI presents the results obtained with a real world dataset to prove the effectiveness of this approach. Fig. 2 (c) shows how the K reference-based approach is used to generate the input vectors for the autoencoder.

## VI. EXPERIMENTS & DISCUSSION

In order to evaluate the proposed autoencoder-based approach for dimension reduction, we ran several tests on a real world gene sequence dataset. The aim of the first set of experiments was to evaluate the proposed approach against a state-of-the-art MDS implementation, and the second set of experiments focused on evaluating the out-of-sample data point performance.

### A. Evaluation Environment

Data visualization is done using WebPlotViz [19],[20], which provides an interactive data visualizer that can plot large 2- and 3-dimensional datasets trivially. DAMDS tests were performed on a cluster which had 16 nodes of Intel Platinum processors with 48 cores in each node, 56Gbps InfiniBand and 10Gbps network connections. Autoencoder-based models were implemented using the distributed data parallel mode of PyTorch [21] and executed on 8 Tesla K80 GPUs.

### B. Data

The dataset used for evaluation is a fungal gene sequence dataset which contains 7 million gene sequences, of which 578K are unique and contain 170K unique sequences that occur more than once within the dataset. The goal of analysing the data is to identify clusters (fungi classes) within the dataset through dimension reduction and visualization. Therefore this dataset is well-suited to evaluate the proposed autoencoder-based dimension reduction approach introduced in this paper. The sequences are RNA sequences made up of "*A,T,G,C*" characters.

The clusters used during the analysis for the dataset were identified through several iterations of pairwise clustering algorithm DAPWC [22], [23], [24] with human input between each iteration. The detailed mechanism used to identify the clusters is beyond the scope of this paper and therefore is not explained in detail. The processes yielded 211 clusters for the 170K unique gene sequences that occur more than once in the gene dataset.

### C. Dimensionality Reduction

In order to compare the dimension reduction capability, we perform dimension reduction of 170K gene sequences to project them in 3-dimensional space. This is done in three approaches which are illustrated in Fig. 2.
- (a) DAMDS will be used as the base to compare against. Input pairwise distance matrix is calculated using a data parallel implementation of Smith-Waterman algorithm. DAMDS implementation is also an MPI-based parallel implementation
- (b) Gene sequences are encoded using OHE to generate the input vectors for the autoencoder. *H* would correspond to *L*C* where *L* is the longest gene sequence in the dataset and *C* is the number of characters in the RNA alphabet. All shorter sequences are padded with zeros to match the length.
- (c) Input vectors for the autoencoder are calculated based on the *K* randomly sampled reference sequences and their similarity based on the Smith-Waterman algorithm.

The first set of experiments were executed on the 170K unique gene sequence dataset. For the autoencoders used for (b) and (c), the autoencoder was structured as "*Inputx128x3x128xOutput*." The activation function used was Relu for all layers other than the bottleneck layer, where LeakyRelu was employed. For (b) the input layer contained 1100 units, which is the *H* value in Fig. 2 (b). For ©, the number of reference sequences *K* was selected as 1500, resulting in an input layer of 1500 units for the autoencoder. Each autoencoder was trained for 100 epochs. The results of the three approaches *a, b, c* are shown in Fig. 3, Fig. 4 and Fig. 5 in order. The visualizations are done in 3-dimensional space, therefore placement of clusters is not completely clear in the figures. Through visual inspection of the 3-dimensional plots it is clear that each method generated an acceptable projection of the data points into 3-dimensional space. However, when compared to DAMDS results, the results of approach (c), which uses reference sequences and Smith-Waterman algorithm, produces better results. Fig. 6 shows distance heat-maps for different runs. They display the correlation between Smith-Waterman distance and the Euclidean distance in the projected 3-dimensional space. The distances are normalized between 0 and 1. The heatmap results show that the approach (c) using reference sequences provides a better mapping than OHE. Also, using 1K references provides a marginally better heatmap than 100 references.

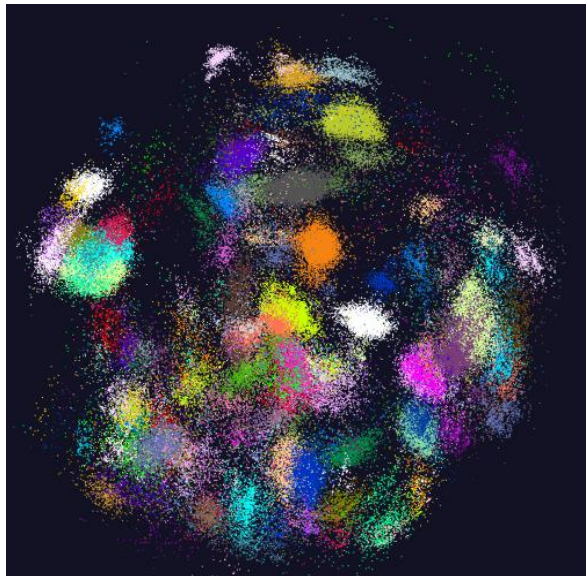

Fig. 3. DAMDS 170K points projected to 3D (a)

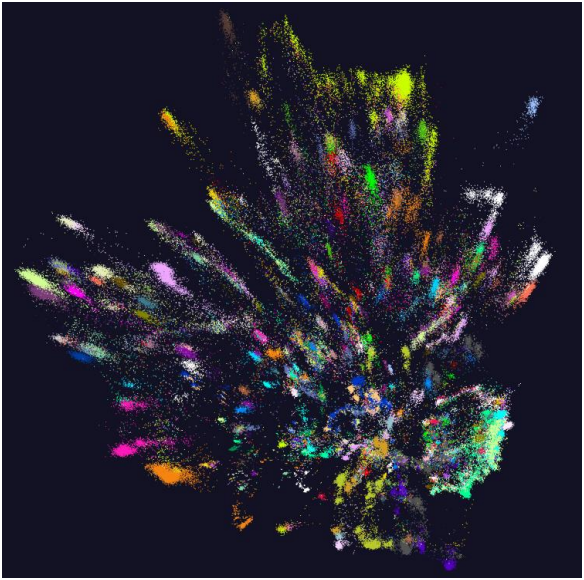

Fig. 4. OHE, 170K points projected to 3D (b)

In order to further evaluate the quality of the dimension reduction achieved through the presented approaches, we used the Silhouette Coefficient (SC) [25] to calculate the quality of the results. SC takes into account both the closeness of points within clusters and the distance between clusters. However, the complex nature of the clusters in the dataset that is projected in 3-dimensional space means SC is not able to capture all the nuances of the clusters. That being said, it still would give some indication on the quality of the results. In order to further analyse approach (c), dimension reduction was performed with varying $K$ to observe the effect of $K$ on the results generated. Table I lists the SC values for each run. Five random reference sequence samples were taken for each $K$ value, and the maximum SC value of the 5 runs are taken. The negative results can be attributed to the complex nature of the clusters in the dataset. The slightly lower SC values for larger $K$ ($>=2000$) can be attributed to the increased network size of the autoencoder to some extent, this effect can be mitigated by increasing the number of layers in the network for larger K as shown in Table III, the results in Table III also suggest that after a certain number of layers the result may go back down. However it is important to note that while SC values show a decline, the visual clusters do not show noticeable degradation. The SC value for DAMDS is comparable to SC values of the reference sequences method when $K$ is sufficiently large. OHE shows lower SC value, which affirms the visual and heatmap results obtained for approach (b).

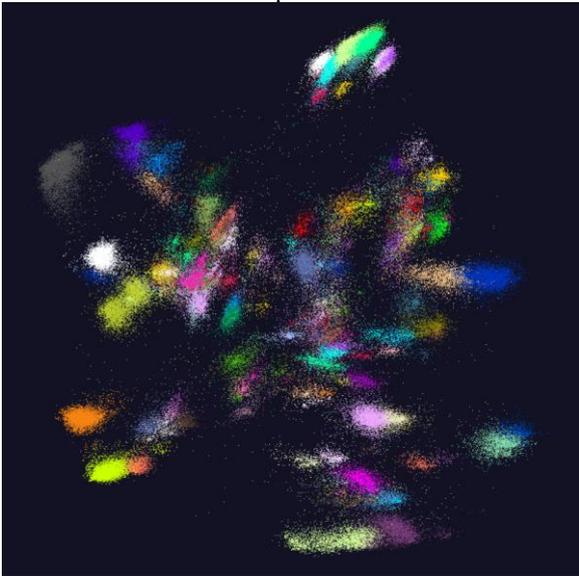

Fig. 5. 1.5K reference sequences, 170K points projected to 3D (c)

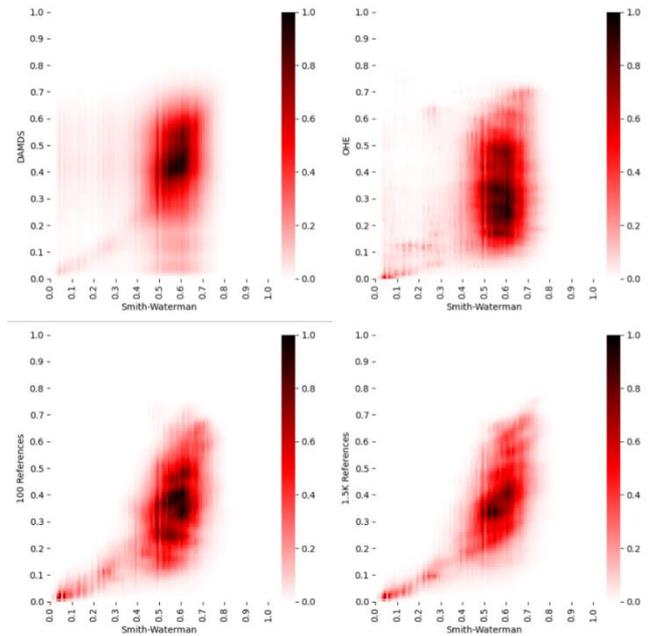

Fig. 6. Heatmaps of Smith-Waterman distance vs projected distance in 3D

### D. Out-of-Sample Data

Evaluating the effectiveness of the autoencoder-based dimension reduction approach for out-of-sample data was performed using a subset of the full 170K data points, which contained 29953 gene sequences with 22 clusters. This was only done for the reference sequence-based method since it showed superior results when compared to the OHE-based method. As the baseline, dimension reduction was performed using $K=400$ with all data points included in the training phase. KMeans algorithm was used to identify the 22 clusters. Afterwards a set of gene sequences were left out for the training phase of the autoencoder but included for the evaluation phase. The resulting 3-dimensional results would again be clustered using KMeans (using the same cluster centers from the baseline results as initialization points). The accuracy for the out-of-sample data points is calculated by counting the number of points which were classified into the same clusters as in the baseline results. For example, if 100 data points were left out during training and 5 of those were classified into a different cluster, then it was classified in the baseline results, which would mean 95% accuracy rate for out-of-sample data. Results listed in Table II show that the trained autoencoder has a very high accuracy for out-of-sample data, more than 99.5%. It is important to note that these accuracy numbers might vary depending on the dataset.

### E. Runtime

Another important aspect that needs to be kept in mind, especially when dealing with large datasets, is the computational complexity and memory requirements of the program in question. Even though availability of resources has grown rapidly over the years, access to large computing clusters to run applications may be hard to come by. One of the main drawbacks of DAMDS is its memory requirement, which is $O(N^2)$. Therefore running large datasets requires a cluster of machines. For example, just to save the distance data matrix for the 170K dataset, roughly 55GB of memory is needed, and the creation of that distance matrix requires roughly 14.5 billion Smith-Waterman calculations as listed in Table IV. If the number of gene sequences in the dataset were to be 1 million, the memory requirement just to load the input dataset would be close to 2TB, and the number of Smith-Waterman calculations would be around 500 billion, which makes it infeasible to run DAMDS on such large datasets. The autoencoder-based solution for dimension reduction is far less demanding on resources. As listed in Table IV, OHE does not have any significant pre-processing requirements, and when using reference sequences, the number of Smith-Waterman distances that need to be calculated are reduced from 14.5 billion to 170 million (for K=1000). Similarly the autoencoder-based method has drastically reduced the memory requirements. In the training and evaluation stage, the memory requirement is mainly dependent on the network size, which in the experiments performed in Section VI-C were only a couple hundred MBs. As the results obtained through the experiments have shown, even a small autoencoder is able to produce accurate outputs. This reduction, especially in memory requirements, allows autoencoder-based dimension reduction scalable to datasets with millions of data points, making dimension reduction on large datasets achievable even with limited resources. While the autoencoder experiments were executed on 8 Tesla K80 nodes, the same network can be trained on a single machine with sufficient memory or a single GPU. We were able to run the same model with 1 Tesla K80 node with ease. While the reduction of computations done during pre-processing is quite clear (14.5 billion to 170 million for 170K sequences), comparing the computation requirements between DAMDS algorithm and the autoencoder-based solution is not straightforward since the two are completely different approaches. However, looking at runtimes for each will provide a sense of the resource requirement differences. It took DAMDS roughly *80* minutes to complete, with a parallelism of 288 on 12 (48 core) compute nodes, while the autoencoder-based solution (K=1000, in Table I) was able to complete within 11 minutes on 8 Tesla K80 GPUs. Even on a single K80 GPU, it was able to complete within 37 minutes, which shows that the autoencoder-based solution was considerably quicker than DAMDS with far less resources.

TABLE I. SILHOUETTE COEFFICIENT FOR DIFFERENT METHODS

| Method | Silhouette Coefficient | | |
|---|---|---|---|
| | *Max* | *Min* | *Average* |
| DAMDS | N/A | N/A | -0.219 |
| One Hot Encoding | N/A | N/A | -0.502 |
| References (K = 25) | -0.233 | -0.388 | -0.309 |
| References (K = 50) | -0.229 | -0.397 | -0.289 |
| References (K = 100) | -0.238 | -0.371 | -0.278 |
| References (K = 200) | -0.236 | -0.331 | -0.268 |
| References (K = 400) | -0.231 | -0.306 | -0.262 |
| References (K = 800) | -0.219 | -0.300 | -0.256 |
| References (K = 1000) | -0.209 | -0.308 | -0.257 |
| References (K = 1500) | -0.193 | -0.311 | -0.248 |
| References (K = 2000) | -0.207 | -0.297 | -0.264 |
| References (K = 4000) | -0.294 | -0.357 | -0.322 |

TABLE II. OUT OF SAMPLE DATA CLASSIFICATION ACCURACY, 400 REFERENCES(K)

| Out-of-Sample Data Points | Incorrect Classification | Accuracy |
|---|---|---|
| 2000 | 4/1000 | 99.8% |
| 4000 | 17/4000 | 99.57% |
| 8000 | 17/8000 | 99.78% |

## VII. RELATED WORK

MDS is a well-researched area and has seen many advances over the years. In [4] the authors use the t-SNE, a variation of Stochastic Neighbor Embedding [26], to visualize high dimensional data in 2- or 3-dimensional space. In [5] the authors present a manifold learning technique named UMAP for dimension reduction that provides better runtime performance than t-SNE. Furthermore, there have been many advances in MDS over the years. [27] provides a survey of many such popular MDS approaches and the real world user cases of each approach.

The use of autoencoders for MDS has also been studied in a limited capacity in the research literature. We believe that this is a promising area which demands further exploration to uncover its full potential. In [16] the authors evaluate the effectiveness of autoencoder-based dimension reduction against other techniques such as PCA and Isomap using both synthetic and real world data. They conclude that in addition to reducing the dimensions, autoencoders can also learn further repetitive patterns in the data which other techniques may not identify. In [28] the authors use deep variational autoencoders for dimension reduction of single cell RNA sequences.

Supporting out-of-sample data for dimension reduction algorithms is an important aspect. While most dimension reduction solutions do not have a direct method to support out-of-sample data, work has been done in this area to provide such capabilities. In [29] the authors introduce a unified framework which offers an extension to support out-of-sample data for several dimension reduction techniques such as Isomap, Local Linear Embedding (LLE), MDS, etc. In [30] the authors present an out-of-sample extension for classical MDS. In [10] the authors discuss an adaptive interpolation method for out-of-sample data points for multidimensional scaling.

TABLE III. SILHOUETTE COEFFICIENT FOR DIFFERENT NETWORK STRUCTURES

| References (K) | | Network (Encoder Hidden Layers) | | | |
|---|---|---|---|---|---|
| | | *128x3* | *256x32x3* | *512x128x32x3* | *1024x256x64x16x3* |
| 4000 | Avg | -0.321 | -0.226 | -0.273 | -0.246 |
| | Max | -0.298 | -0.180 | -0.210 | -0.202 |
| | Min | -0.349 | -0.260 | -0.331 | -0.288 |
| 6000 | Avg | -0.356 | -0.205 | -0.239 | -0.233 |
| | Max | -0.258 | -0.188 | -0.174 | -0.200 |
| | Min | -0.482 | -0.218 | -0.307 | -0.261 |

TABLE IV. COMPUTATIONS INVOLVED IN EACH APPROACH, FOR 170K FUNGI GENE DATASET

| Method | Pre-Processing Stage | Dimension Reduction Stage |
|---|---|---|
| DAMDS | SW Distance Calculations to generate input distance matrix (~14.5 billion calculations). SW algorithm is $O(ML)$ where $M$ and $L$ are lengths of each gene sequence & DAMDS algorithm - $O(N^2)$ | DAMDS algorithm - $O(N^2)$ Runtime with 288 way parallelism on 12 (48 core) nodes ~80 minutes |
| OHE | N (170K) one hot encoding calculations to generate input vectors. OHE has a complexity of $O(N)$ | Autoencoder training phase, the runtime calculations depend on the number of epochs and network complexity. Add evaluation phase to generate the dimension-reduced results. Runtime on 1 Tesla K80 GPU ~37 minutes. Runtime with 8 way parallelism on 8 Tesla K80 GPUs ~11 minutes. |
| Reference Sequences | Smith-Waterman Distance Calculations to generate input distance matrix (~170 million calculations for *K=1000*). SW algorithm is $O(ML)$ where $M$ and $L$ are lengths of each gene sequence. | Autoencoder training phase; the runtime calculations depend on the number of epochs and network complexity. Add the Evaluation phase to generate the dimension-reduced results. Runtime on 1 Tesla K80 GPU ~37 minutes. Runtime with 8 way parallelism on 8 Tesla K80 GPUs ~11 minutes. |

## VIII. CONCLUSIONS

This paper presented and evaluated an autoencoder-based dimension reduction approach for large gene sequence datasets. From the results that we obtained from several experiments on real gene sequence data, and comparing them with results obtained from DAMDS, it is clear that the autoencoder-based solution which uses a set of reference sequences is capable of producing comparable (or better) results to DAMDS while drastically reducing resource and computation requirements. The results also showcase how accurate the model is for out-of-sample data points, which is directly supported without any additional work needed in the presented model. Therefore we can conclude that the autoencoder-based dimension reduction approach coupled with reference sequence method for input encoding is a viable and promising approach to analyse large gene sequence datasets, with high accuracy and minimal compute resource requirements, making such data analysis tasks easily accessible to a broader audience.

While this paper focused on the evaluation of the model around gene sequence data, it is plausible that this model can be applied to many other forms of data. However this would require further research and experiments to verify the model accuracy and validity. A strong argument in favour of the applicability onto a broader range of datasets would be the well-known versatility of neural networks that has been observed throughout the past several years in the research literature. One direct extension would be to apply such a model for large biological image datasets by adding convolution layers to the network. This is left as future work alongside the evaluation of the autoencoder-based models for various types of datasets in pursuit of dimension reduction.

ACKNOWLEDGMENT

This work was partially supported by NSF CIF21 DIBBS 1443054, CINES 1835598 and Global Pervasive Computational Epidemiology 1918626, DoE DE-SC0021418 and the Indiana University Precision Health initiative. We thank Intel for their support of the Juliet and Victor systems and extend our gratitude to the FutureSystems team for their support with the infrastructure. We would also like to extend our thanks to Anna Rosling from Uppsala University, Sweden for the fungi gene sequence dataset used for experiments.